%% file: emnlp2023.tex
\pdfoutput=1

\documentclass[11pt]{article}

\usepackage[]{EMNLP2023}

\usepackage{times}
\usepackage{latexsym}
\usepackage{multirow}
\usepackage{hyperref}
\usepackage[T1]{fontenc}

\usepackage[utf8]{inputenc}

\usepackage{microtype}

\usepackage{booktabs}
\usepackage{float}
\usepackage[normalem]{ulem}

\usepackage{inconsolata}

\usepackage{graphicx}

\author{
    Fardin Ahsan Sakib,
    A H M Rezaul Karim,
    Saadat Hasan Khan,
    Md Mushfiqur Rahman\\
    Department of Computer Science, George Mason University 
    \footnote{All authors have made equal contributions}\\
    \texttt{\{fsakib,akarim9,skhan225,mrahma45\}@gmu.edu} 
}
    
\title{Intent Detection and Slot Filling for Home Assistants: Dataset and Analysis for Bangla and Sylheti}
\begin{document}
\maketitle

\begin{abstract}

As voice assistants cement their place in our technologically advanced society, there remains a need to cater to the diverse linguistic landscape, including colloquial forms of low-resource languages. Our study introduces the first-ever comprehensive dataset for intent detection and slot filling in formal Bangla, colloquial Bangla, and Sylheti languages, totaling 984 samples across 10 unique intents. Our analysis reveals the robustness of large language models for tackling downstream tasks with inadequate data. The GPT-3.5 model achieves an impressive F1 score of 0.94 in intent detection and 0.51 in slot filling for colloquial Bangla. 
\footnote{The dataset and the analysis code can be found in the following directory: \href{https://github.com/mushfiqur11/bangla-sylheti-snips.git}{https://github.com/mushfiqur11/bangla-sylheti-snips.git}}



\end{abstract}

\input{Sections/intro.tex}
\input{Sections/dataset}
\input{Sections/method.tex}
\input{Sections/experiment}
\input{Sections/results}
\input{Sections/conclusion.tex}
\input{Sections/limitations}

\bibliography{Bib/anthology,Bib/custom}
\bibliographystyle{Bib/acl_natbib}

\appendix

\input{Appendix/Appendix}



\end{document}

%% file: Sections/intro.tex
\section{Introduction}
Smart devices have become commonplace, establishing home assistants as indispensable fixtures in contemporary households. These voice-activated virtual companions adeptly manage an array of tasks, ranging from setting reminders to controlling room temperatures. The efficacy of home assistants in performing these tasks is closely intertwined with their underlying Natural Language Understanding (NLU) models, which enable seamless interactions in high-resource languages \cite{chen2019bert, stoica2021intent, antoun2020arabert, upadhyay2018almost}. However, this advantage in NLU capabilities is not extended to low-resource languages \cite{stoica2019impact, schuster2018cross}, presenting a notable discrepancy. This discrepancy holds considerable significance, especially considering the global demand for home assistants and the extensive usage of low-resource languages, which have a substantial speaker base.

\par Bangla and Sylheti \cite{ethnologue200}, with 285 million native speakers combined, have rich cultural and colloquial nuances. Specialized datasets are needed to capture these intricacies as users prefer to interact with home assistants in their native languages, highlighting the research need \cite{bali2019ellora}.


\par The language understanding of home assistants is dependent on two key NLU tasks: intent detection and slot filling \cite{weld2022survey, louvan2020recent}. Intent detection determines user actions, like playing music or checking the weather, while slot filling extracts specific details, such as song titles or locations. These tasks enable seamless human-device interactions, especially for home assistants.

\par Research on intent detection and slot filling primarily focuses on high-resource languages \cite{liu2016attention, qin2021co, niu2019novel, zhang2018joint}. While there have been limited studies dedicated to the Bangla language \cite{bhattacharjee2021banglabert, alam2021review, hossain2020banfakenews}, none of them have addressed the tasks of intent detection and slot filling in Bangla. Furthermore, these studies have not taken into account colloquial variants or closely related languages like Sylheti. This gap in research leaves a significant portion of the speaker base underserved.

\par This paper bridges this research gap with several notable contributions. Firstly, we introduce a comprehensive dataset encompassing 328 entries for intent detection and slot filling for each of the three languages – totaling 984 samples. These languages include formal Bangla, colloquial Bangla, and colloquial Sylheti. We further show a comparative study between generative LLMs and state-of-the-art language models for intent detection and slot filling.

%% file: Sections/dataset.tex
\section{Dataset}
\par At the core of our exploration stands a meticulously curated dataset that is inspired by the SNIPS dataset \cite{coucke2018snips}, which caters to the broad audience. 

\subsection{Dataset Size and Distribution}

\par Originating from the 328 English samples present in the SNIPS dataset, our dataset underwent a manual correction phase to ensure that the English samples were of optimal quality. Then, we created three linguistically diverse variants, maintaining the same distribution across intent classes and slots as the original samples. These are:
\begin{enumerate}
    \item\textbf{Formal Bangla:} This represents the standard version of the Bangla language, majorly used in contexts like official documents, news broadcasts, and literature. Formal Bangla tends to adhere strictly to grammatical rules.
    \item\textbf{Colloquial Bangla:} An informal variant predominantly used in Bangladesh, colloquial Bangla resonates with everyday conversations of its people. While there are numerous dialects in different regions of Bangladesh, this form remains more or less consistent across the country. Colloquial Bangla is more flexible regarding syntax and incorporates a significant number of loanwords from English, Arabic, Persian, and other languages.
    \item\textbf{Colloquial Sylheti:} A language with unique intricacies, Sylheti stands apart from Bangla and is spoken in the Sylhet region of Bangladesh and among diaspora communities. It's rich in expressions, proverbs, and idiomatic language that reflect the history and culture of the Sylhet region.
\end{enumerate}

\par The curated dataset spans 10 distinctive intents. Each specific intent has a distinct set of slot categories. Figure \ref{sample_per_intent} shows the number of samples for each intent and Figure \ref{slot_cat} shows the fraction of slots that frequently occur for each intent, with respect to infrequently occurring slots.


\subsection{Data Generation Process}

The generation of our dataset was methodical and rigorous to ensure authenticity and accuracy.\\
\noindent \textbf{Annotator Engagement} \\
Four doctoral students were on board as annotators for our project. The initial phase involving the rectification of English data from the SNIPS dataset was a collaborative effort, with each annotator working on a distinct, non-overlapping segment. Subsequent phases involved two individuals fluent in Bangla for the Bangla datasets and two native Sylheti speakers for the colloquial Sylheti dataset.\\
\noindent \textbf{Base Creation} \\
The base dataset was created using the Bangla-T5 model \cite{bhattacharjee2023banglanlg}, a state-of-the-art English-to-Bangla translation tool, following the work of \citeauthor{de-bruyn-etal-2022-machine}. The refined English samples served as the foundation to produce the initial Bangla translations for each sample. An auto-generated dataset comes with a myriad of issues. Therefore, these samples were manually re-translated and annotated with the auto-translations as the base.\\
\noindent \textbf{Inter-Annotator Agreement} \\
An essential step in ensuring the reliability of our dataset was to gauge the consistency between annotators. For each language variant, 28 randomly chosen samples were annotated independently by both designated annotators, followed by calculating their inter-annotator agreement (Table \ref{annotator-agreement}). This exercise helped us discern the degree of concordance and areas of divergence.\\
\noindent \textbf{Consensus Building} \\
Post the initial agreement calculation, a meeting was convened where the annotators discussed and reconciled their differences. This step was instrumental in ironing out inconsistencies and ensuring a unified approach going forward.\\
\noindent \textbf{Blind Overlap} \\
As the annotators progressed with data creation, a random 10\% of the samples were earmarked for blind overlap. These served as a secondary check on inter-annotator agreement after dataset creation.\\
\noindent \textbf{Independent Adjudication} \\
After the final compilation of the dataset, each entry underwent a rigorous review by an independent adjudicator who had not previously worked on that particular language variant. This added an additional layer of scrutiny and quality assurance.

\begin{table}[ht!]
\centering
\small
\begin{tabular}{ccc}
\toprule
\multicolumn{3}{c}{Inter-annotator agreement} \\
\midrule
 & \begin{tabular}[c]{@{}c@{}}Cohen's\\ Kappa\end{tabular} & \begin{tabular}[c]{@{}c@{}}Average\\ BLEU\end{tabular} \\
 \midrule
First 28 samples & 0.42 & 0.43 \\
Blind overlap (10\%) & 0.55 & 0.51 \\
\bottomrule
\end{tabular}
\caption{There was an increase in annotator agreement before and after the annotator's meeting. This ensures the homogeneity of annotations in the dataset.}
\label{annotator-agreement}
\end{table}

\subsection{Ensuring Quality}
\begin{figure}[t]
    \centering
    \includegraphics[width=0.82\linewidth]{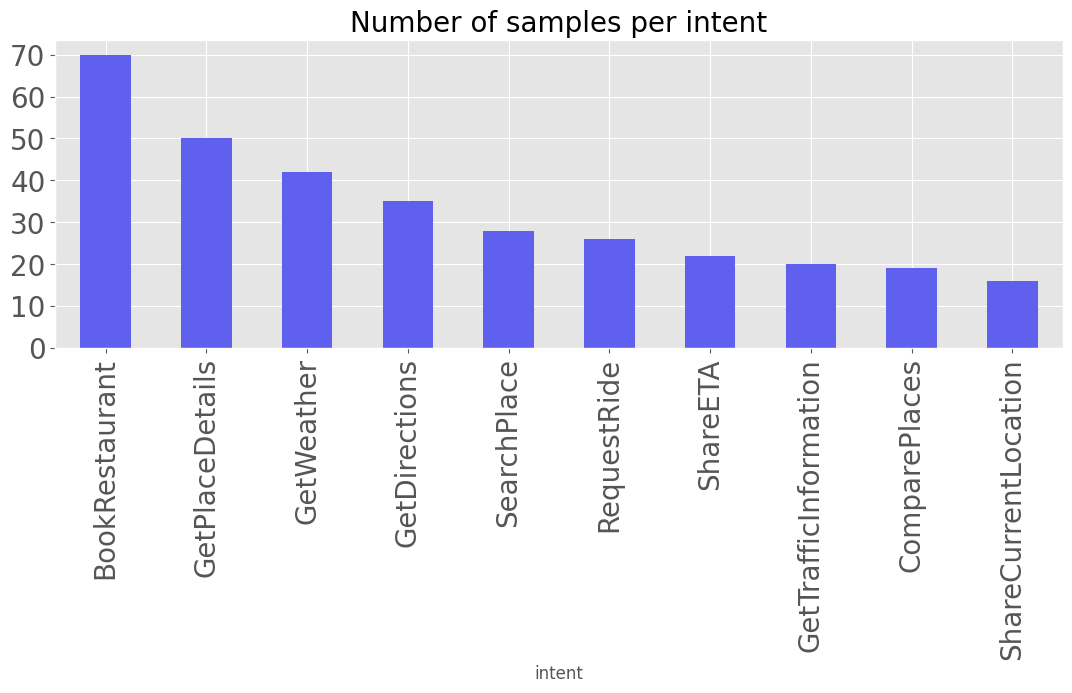}
    \caption{The number of samples for each intent varies, but they are fairly distributed, with 18 to 68 samples per intent.}
    \label{sample_per_intent}
\end{figure}

\begin{figure}[t]
    \centering
    \includegraphics[width=0.82\linewidth]{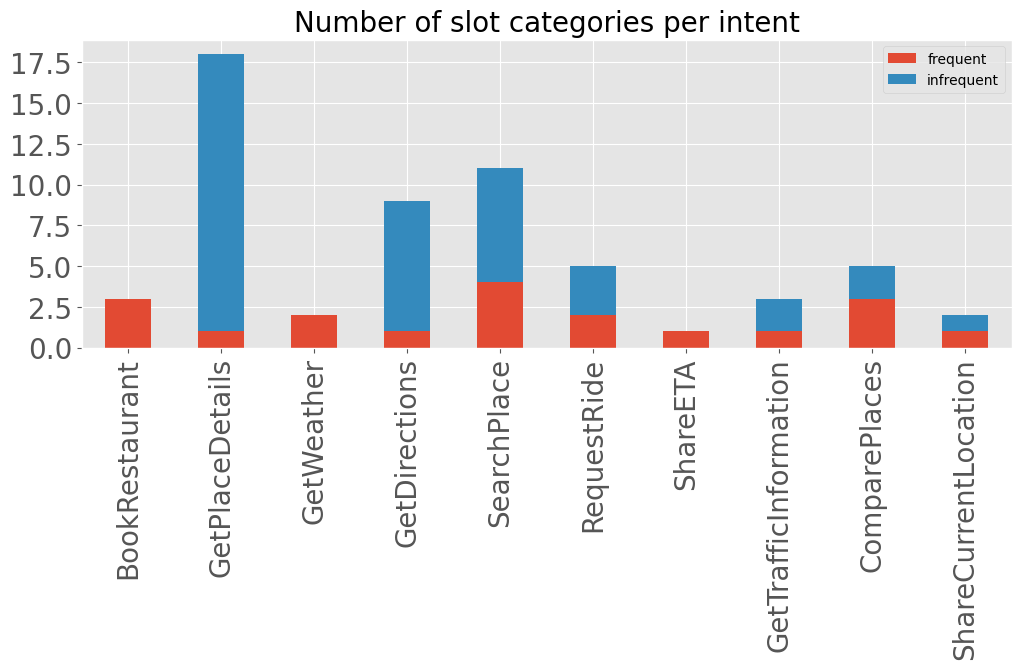}
    \caption{Slot categories appearing in at least 30\% of the instances are marked as "frequent," while others are "infrequent." Despite varying slot categories per intent, frequent ones are evenly distributed.}
    \label{slot_cat}
\end{figure}

Our data generation process, featuring multiple checks, blind overlaps, third-party reviews, and inter-annotator agreement stages, highlights our commitment to quality. It minimizes biases and discrepancies that could result from a single annotator's viewpoint. The inclusion of an independent adjudicator in the final review further bolsters the dataset's integrity and reliability. Using a well-established dataset as the baseline ensures proper distribution of the data across different labels (Figure \ref{sample_per_intent} and Figure \ref{slot_cat}). 


%% file: Sections/method.tex
\section{Methodology and Experimental Setup}

Our experiments were divided into four phases. In our initial experiment, we employed JointBERT \cite{chen2019bert}, the state-of-the-art model in this domain, for both intent detection and slot-filling tasks. In our next experiment, JointBERT was retained for intent detection, while we explored the capabilities of GPT-3.5 (Generative Pre-trained Transformer) \cite{brown2020language} model for slot filling. The third experiment fully utilized GPT-3.5 for both tasks. For our concluding experiment, we provided GPT-3.5 with the original intents and then analyzed its performance on the slot-filling task. The final experiment gives the raw result of slot-filling for the GPT model.
\par \textbf{JointBERT} leverages the BERT \cite{devlin2019bert} model to provide a unified approach encompassing both intent classification and slot filling by utilizing the representations from the pre-trained BERT model. We employed the default BERT tokenizer and maintained consistent parameters for all three languages. The utilization of these default settings and tokenization methods ensures an equitable and consistent evaluation across the languages.
\par
\begin{figure}[ht!]
    \centering
    \includegraphics[width=0.85\linewidth]{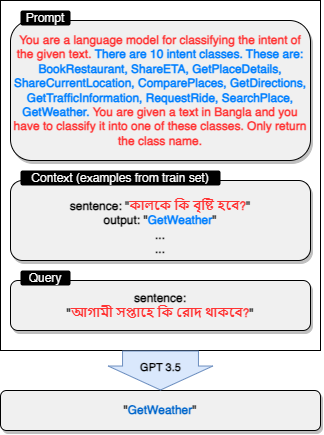}
    \caption{The figure illustrates how the input is formatted for the intent-detection task. A base-prompt is passed on to the GPT model. A few samples (5) from the training set are also passed as the context. From these sentence-output pairs, the LLM understands how the task needs to be solved. Finally, the current query is passed}
    \label{fig:intentprompt1}
\end{figure}

\begin{figure}[ht!]
    \centering
    \includegraphics[width=0.85\linewidth]{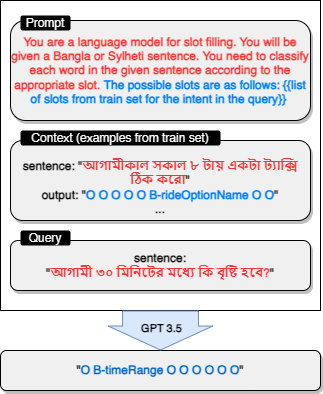}
    \caption{The input structure for the slot-filling task is quite similar to the intent detection task. The major difference is the prompt. For slot-filling, the set of possible slots is based on the intent type of the query. The intent type is obtained from a separate model and then from the train set, all possible slots for the given intent are fetched}
    \label{fig:slotprompt}
\end{figure}

 \textbf{GPT-3.5} (Generative Pre-trained Transformer) \cite{brown2020language} model operates on the Transformer architecture and is adept at generating text resembling human language by predicting subsequent words or tokens in a sequence. GPT-3.5's deep contextual understanding is a result of extensive pre-training on a diverse corpus of textual data, encompassing various languages and linguistic intricacies enabling it to excel across a spectrum of NLP tasks \cite{goyal2022news,liu2021makes, sakib2023extending,kumar2020data}. We used GPT in a few-shot setting, passing 5 training samples along with the prompt. Rigorous prompt engineering was performed before settling on the two prompts for the two tasks. Figure \ref{fig:intentprompt1} and Figure \ref{fig:slotprompt} show the final versions of the prompts used in the experimentations.

\subsection{Experimental Setup}

We divided each of the three datasets into training, development, and test sets using a standard 80-10-10 split. The JointBERT model was trained and evaluated on an A100 GPU, using a batch size of 8. We closely followed the setup provided by the original authors for this phase. For GPT, we used the OPENAI API with the ``GPT-3.5-turbo'' engine and set the token limit to 50. 


%% file: Sections/experiment.tex


%% file: Sections/results.tex
\section{Results}

Tables \ref{intent-table} and \ref{slot-table} present the performance of the models we evaluated on our intent detection and slot-filling tasks. A clear pattern emerges: GPT-3.5 consistently outperforms JointBERT in both tasks.

While intent detection is generally more straightforward, JointBERT performs reasonably well in this aspect, although it doesn't quite match the exceptional performance achieved by GPT-3.5. However, when it comes to the more intricate task of slot-filling, JointBERT's performance falls significantly short, leaving ample room for improvement. In contrast, GPT-3.5 demonstrates its proficiency in handling the complexities of this task.

A significant reason behind GPT-3.5's superior performance is its broader exposure to diverse languages during training, including Bangla. JointBERT, conversely, hasn't been specifically trained on any Bangla dataset. This linguistic familiarity gives GPT-3.5 a clear advantage, enabling it to process and interpret Bangla's nuances far more effectively than JointBERT. The results underline the significance of using LLMs for low-resource languages, especially in scenarios where obtaining high volumes of training data for a particular downstream task is challenging.

\begin{table}[t]
\small
\centering
\begin{tabular}{c|ccc}
\toprule
         \multicolumn{4}{c}{Intent Detection (\textit{Accuracy and F1 Score})}                                                                                                                                                             \\
\midrule
Models   & \begin{tabular}[c]{@{}c@{}}Formal\\ Bangla\end{tabular} & \begin{tabular}[c]{@{}c@{}}Colloquial\\ Bangla\end{tabular} & \multicolumn{1}{c}{\begin{tabular}[c]{@{}c@{}}Colloquial\\ Sylheti\end{tabular}} \\
\midrule
JointBERT & 0.57 | 0.56                                                   & 0.63   | 0.61                                                     & 0.45 | 0.46                                                         \\
GPT-3.5  & 0.94 | 0.94                                                   & 0.94   | 0.94                                                     & 0.87 | 0.89        \\                                               \bottomrule
\end{tabular}

\caption{While the performance of JointBERT is noteworthy for Bangla and its variants, the GPT-3.5 model excels across all metrics for all three datasets}
\label{intent-table}
\end{table}

\begin{table}[t]
\centering

\small
\setlength{\tabcolsep}{3pt}
\begin{tabular}{cc|ccc}
\toprule
\multicolumn{5}{c}{Slot Filling (\textit{F1 Score})} \\
\midrule
\begin{tabular}[c]{@{}c@{}}Slot Filling\\ Model\end{tabular} & \begin{tabular}[c]{@{}c@{}}Intent\\ From\end{tabular} & \begin{tabular}[c]{@{}c@{}}Formal\\ Bangla\end{tabular} & \begin{tabular}[c]{@{}c@{}}Colloquial\\ Bangla\end{tabular} & \begin{tabular}[c]{@{}c@{}}Colloquial\\ Sylheti\end{tabular} \\
\midrule
JointBERT & JointBERT & 0.14 & 0.11 & 0.07 \\
GPT-3.5 & JointBERT & 0.43 & 0.45 & 0.52 \\
GPT-3.5 & GPT-3.5 & 0.45 & 0.51 & 0.57 \\
GPT-3.5 & Original & 0.54 & 0.53 & 0.57 \\
\bottomrule
\end{tabular}

\caption{ The slot-filling task is separate from but dependent on the intent detection task. Intent needs to be passed to the model for good performance. In slot-filling tasks, GPT massively outperforms JointBERT}
\label{slot-table}
\end{table}

%% file: Sections/conclusion.tex
\section{Conclusion}

In the era of smart devices, a home assistant's voice interfaces must resonate with the authentic linguistic intricacies of its users. Our research presents the first-ever dataset for intent detection and slot filling in Bangla and Sylheti, emphasizing their colloquial forms. This focus on colloquial forms bridges the often-overlooked gap between formal language models and the nuances of everyday speech. By championing colloquial forms, we ensure a voice interface that's more natural and attuned to genuine communication habits. Through rigorous data collection and validation, we have produced a high-quality benchmark dataset, providing a solid foundation for subsequent analyses and model evaluations. The comparative study between large language models (LLM) like GPT-3.5 and non-LLMs underscores the remarkable capability of LLMs to excel even with minimal datasets, marking a considerable stride for underrepresented languages. 


%% file: Sections/limitations.tex
\section{Limitations}

While our research has made significant strides in understanding intent detection and slot filling for Bangla and Sylheti, like any study, it has its limitations. Our dataset, although carefully curated for the Bangla and Sylheti variants, is on the smaller side compared to established benchmarks. A precise and robust data generation process was prioritized, naturally limiting our data volume. We confined our evaluations to the JointBERT model and GPT-3.5. The pronounced difference in their performance deterred us from testing a broader range of models. Moreover, the dearth of optimized Bangla models for specific tasks posed challenges. An attempt with a Bangla BERT tokenizer didn’t yield satisfactory outcomes, affecting the JointBERT's efficacy. As promising as our results are, they are tied to our specific dataset and context. Extending our findings to diverse settings or other languages requires further exploration, marking just the beginning of this exciting journey.

%% file: Appendix/Appendix.tex
\section{Appendix}

\subsection{Related Work}
 
\par Efforts to enhance datasets for intent detection and slot-filling within low-resource languages, such as Bangla and Sylheti in this context, commence with the intricate process of translating individual English lexemes extracted from established benchmarks like ATIS and SNIPS. Previous works in intent detection and slot filling for low resource languages\cite{dao2021intent, akbari2023persian}, have translated each English utterance to their respective languages. Recent works have shown that there are great performance achievements on intent detection and slot-filling tasks on datasets that have been derived from the SNIPS dataset \cite{weld2022survey, qin2019stack, wang2020encoding}, and this gives a reason to choose the SNIPS dataset over the ATIS dataset as it is a good starting point for a work with a language that has never been explored.


\noindent \par Spoken Language Understanding, a pivotal endeavor in the domain of task-oriented dialogue systems, encompasses the tasks of intent detection and slot-filling. Traditionally, these tasks were regarded as distinct domains in which significant progress was made \cite{tur2012towards,ravuri2015recurrent,mesnil2013investigation,vu2016bi}. However, recent research has garnered notable attention by achieving remarkable advancements in performance through the concurrent learning of intent detection and slot-filling tasks \cite{zhang2018joint, weld2022survey}. In this section, we're primarily looking at how intent detection and slot-filling tasks are combined. We'll focus on two well-known strategies for this integration:

\begin{itemize}
    \item A strategy devised through parameter sharing and the exchange of hidden states, utilizing a common BiLSTM/BERT encoder, along with two distinct decoders dedicated to intent detection and slot filling, on top of the shared encoder. \cite{chen2019bert,xu2013convolutional,liu2016attention,zhang2016joint}.
    \item Another strategy, extending the initial approach to a more advanced level, involves the model acquiring an understanding of the relationships between slots and intent labels. This frontier has been explored in research in two distinct ways. Some studies \cite{goo2018slot,li2018self,niu2019novel} have demonstrated the use of attention mechanisms to discern the correlation between the overarching intent context representation and the slot vectors generated by the encoder. Alternatively, other works \cite{qin2019stack,zhang2019joint} have approached this by initially learning the representation of the utterance, which aligns with the representation of the global intent context, utilizing a self-attention mechanism. Subsequently, they join this representation with the encoder's vector outputs before feeding the combined vectors into the slot-filling decoder.

\end{itemize}

\subsection{Examples from the dataset}
Here we include a few examples from each of the datasets. 

\begin{figure}[!ht]
    \centering
    \includegraphics[width=0.8\linewidth]{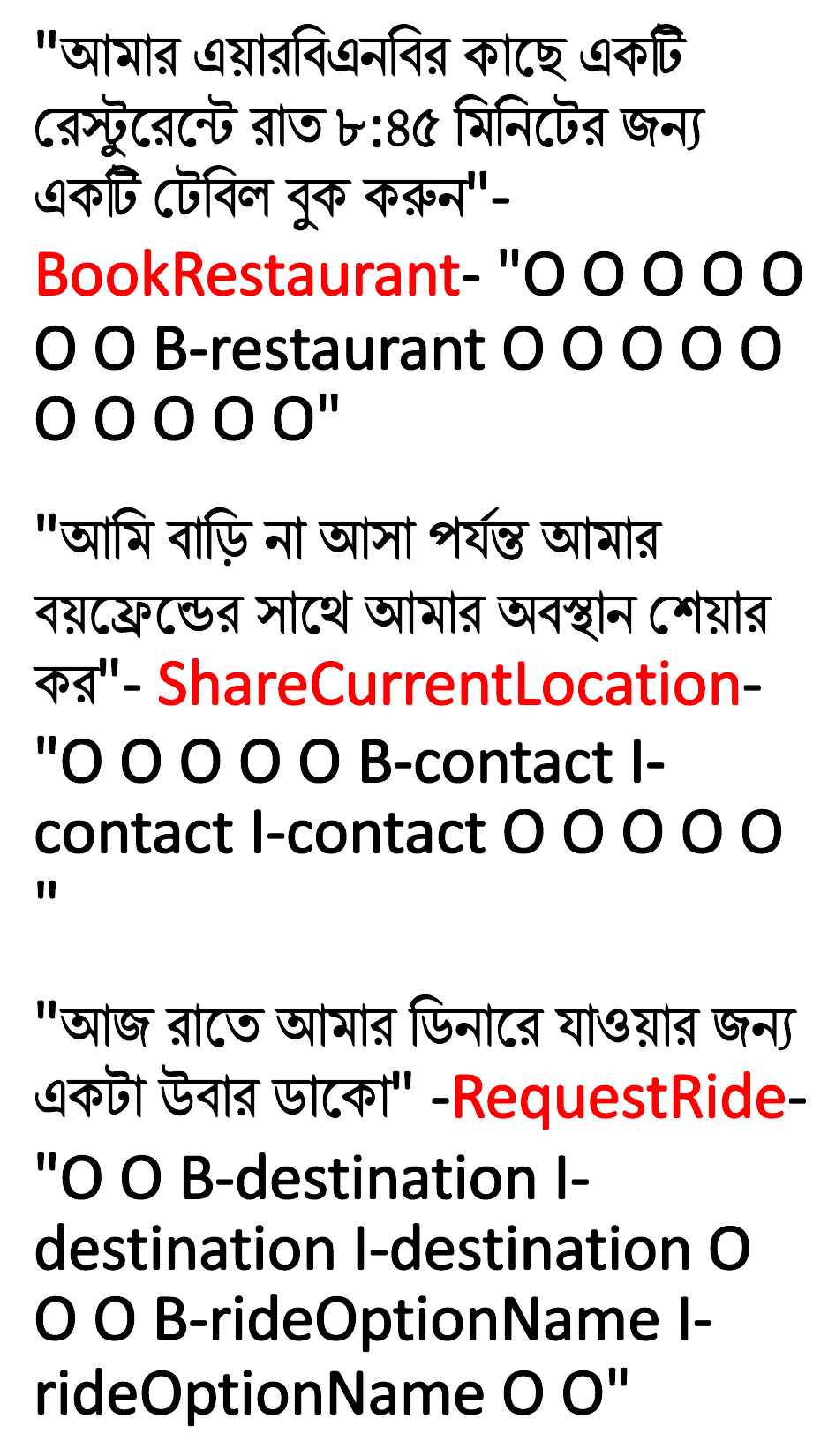}
    \label{fig:intentprompt2}
    \caption{Few examples from the Formal Bangla dataset. (Input sentence - Intent - Expected slots)}
\end{figure}

\begin{figure}[!ht]
    \centering
    \includegraphics[width=0.8\linewidth]{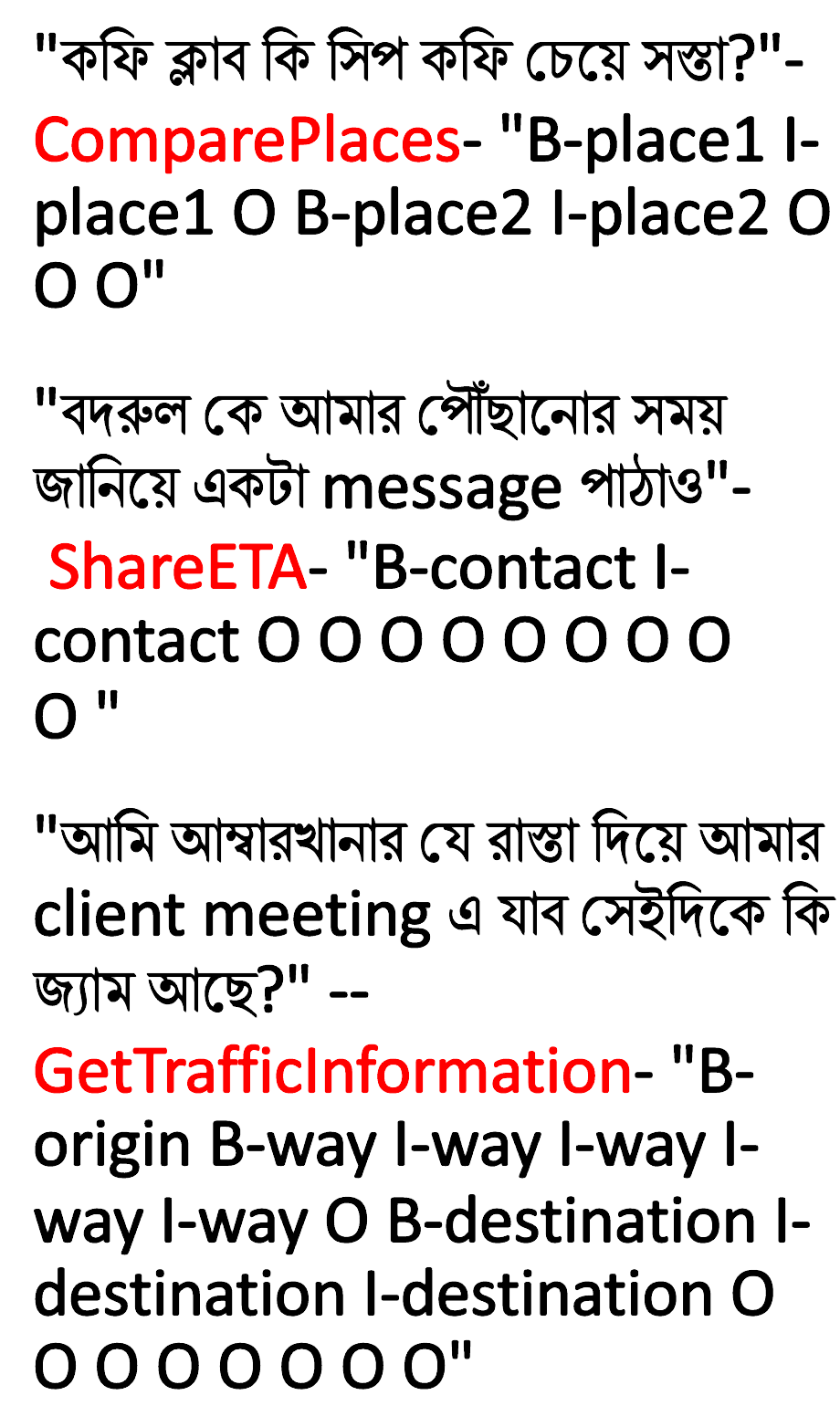}
    \label{fig:intentprompt3}
    \caption{Few examples from the Colloquial Bangla dataset. (Input sentence - Intent - Expected slots)}
\end{figure}

\begin{figure}[!ht]
    \centering
    \includegraphics[width=0.8\linewidth]{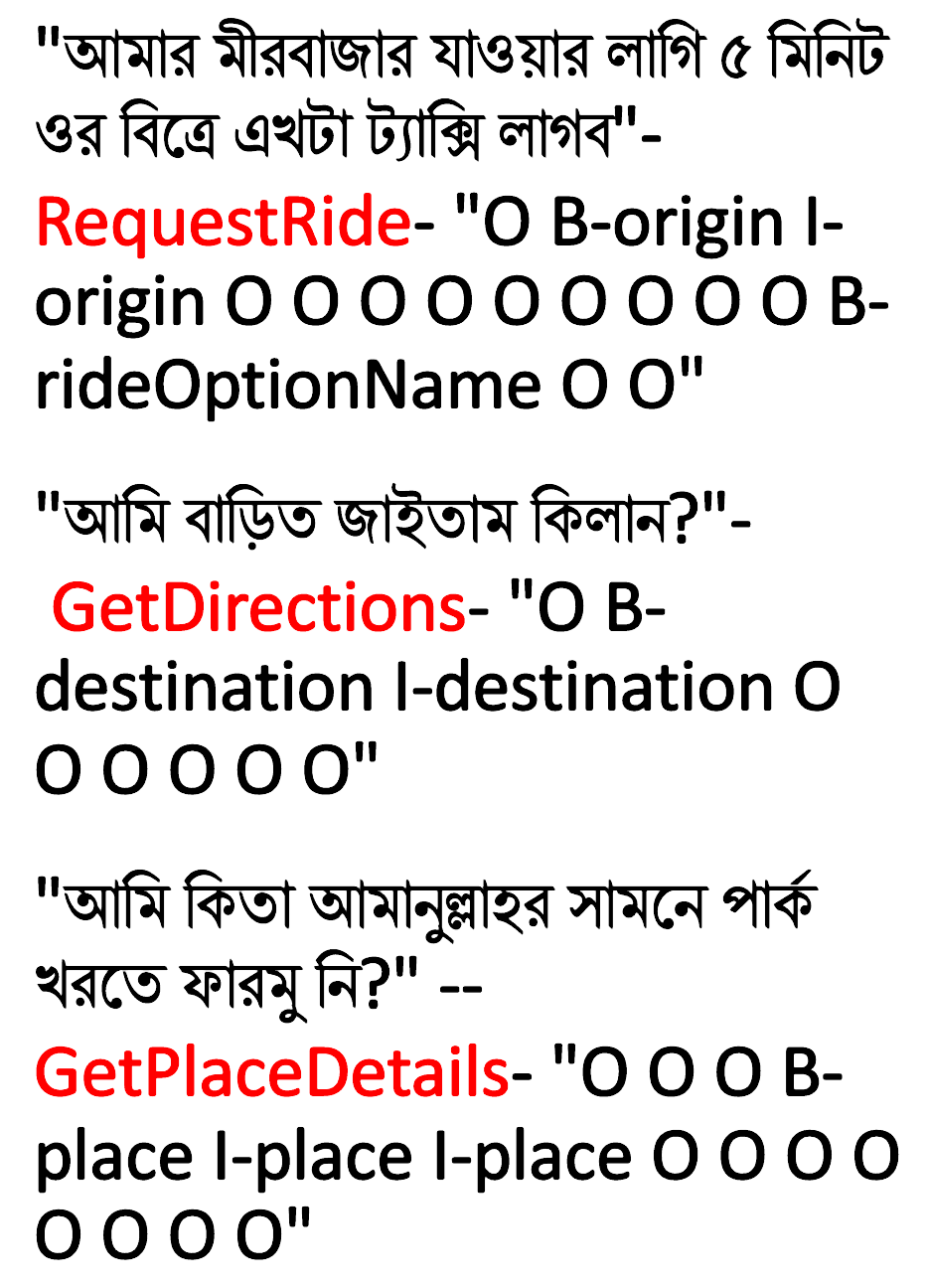}
    \label{fig:intentprompt4}
    \caption{Few examples from the Sylheti dataset. (Input sentence - Intent - Expected slots)}
\end{figure}

\subsection{Prompts used for GPT}
For the intent detection task we used the following prompt:
\textit{"You are a language model for classifying the intent of the given text. There are 10 intent classes. These are: BookRestaurant, ShareETA, GetPlaceDetails, ShareCurrentLocation, ComparePlaces, GetDirections, GetTrafficInformation, RequestRide, SearchPlace, GetWeather. You are given a text in Bangla and you have to classify it into one of these classes. Only return the class name."}

In this approach, we clearly outlined the potential intent classes, specified the input language as Bangla, and directed the model to solely return the class name. Such structuring was essential to elicit precise responses from the model.

For our slot-filling task, we utilized the following prompt:
\textit{"You are a language model for slot filling. You will be given a Bangla sentence. You need to classify each word in the given sentence according to the appropriate slot. The possible slots are as follows: {{list of possible slots extracted from the train set (based on the training intent)}}"}

We equipped the model with both the potential slots and their associated intent. Notably, the performance fluctuated depending on the source of the intent— GPT-3.5, JointBERT, or the Original dataset.

\subsection{Inter-annotator metrics}

In order to assess inter-annotator agreement, this study utilized two primary evaluation metrics: Cohen's Kappa and Average BLEU.

Cohen's Kappa provides a statistical measure of agreement between two annotators, while accounting for the possibility of chance agreement. Specifically, it involves calculating the actual observed agreement between the annotators and comparing that to the level of agreement that would be expected by random chance. Cohen's Kappa expresses the ratio between these two values as a score ranging from 0 to 1, with higher scores indicating greater reliability.

Average BLEU (Bilingual Evaluation Understudy) is a commonly employed metric for evaluating machine translation outputs by comparing them against one or more reference translations. It analyzes the co-occurrence of n-grams between the translated text and human reference texts to produce a score reflecting the quality and fluency of the translation. Taking the average BLEU score across multiple translations provides an overall indicator of the fidelity of the translations with respect to the reference materials.

Together, these two metrics enable analysis of both the reliability of individual annotators via Cohen's Kappa and the accuracy and fluency of translations via Average BLEU in relation to trusted references. The combination provides a robust means of evaluating key aspects of annotation quality for this study.